\documentclass{INTERSPEECH2023}
\interspeechcameraready 
\usepackage{color}
\usepackage[T1]{fontenc}
\usepackage{svg}
\usepackage[disable]{todonotes}

\title{Improving Isochronous Machine Translation \\with Target Factors and Auxiliary Counters}
\name{Proyag Pal$^{1,2}$*\thanks{*This work was done during an internship at Amazon.}, Brian Thompson$^1$, Yogesh Virkar$^1$, Prashant Mathur$^1$, \\ Alexandra Chronopoulou$^{1,3}$*\footnotemark[1], Marcello Federico$^1$}
\address{
  $^1$AWS AI Labs, USA \\
  $^2$School of Informatics, University of Edinburgh, Scotland\\
  $^3$Center for Information and Language Processing, LMU Munich, Germany
}
\email{brianjt@amazon.com}

\begin{document}

\maketitle

\begin{abstract}
To translate speech for automatic dubbing, machine translation needs to be isochronous, i.e. translated speech needs to be aligned with the source in terms of speech durations. We introduce target factors in a transformer model to predict durations jointly with target language phoneme sequences. We also introduce auxiliary counters to help the decoder to keep track of the timing information while generating target phonemes.  We show that our model improves translation quality and isochrony compared to previous work where the translation model is instead trained to predict interleaved sequences of phonemes and durations.
\end{abstract}

\section{Introduction}
Automatic dubbing \cite{federico-etal-2020-speech} aims to translate speech from video content (such as movies and TV shows) into a target language while maintaining isochrony, i.e. matching the speech and pause structure of the source speech in order to preserve time synchronization in the dubbed video. 
In the standard automatic dubbing pipeline, an automatic speech recognition (ASR) system transcribes the source audio into source language text, the text is translated into the target language by a machine translation (MT) system, after which a prosodic alignment (PA) module inserts pauses to segment the translated text, before a text-to-speech (TTS) system generates target language speech.

One drawback of this pipeline is the fact that since the machine translation system is unaware of isochrony constraints, it can generate translations which do not fit the timing of the source audio.
After segmenting target text through the PA module, to ensure the segments fit the speech timing, the speaking rate has to be adjusted for the TTS system, often resulting in unnatural output speech.

Our goal is to jointly optimize translation quality and isochrony, i.e. predict translations and target-side timing information using the same model to generate translations of high quality while matching the source's speech timing. We achieve this using target factors \cite{garcia-martinez-etal-2016-factored}, 
where alongside predicting phoneme sequences as the target, we also predict durations for each phoneme as a target factor. Additionally, we design auxiliary counters\footnote{The counters are modified target factors providing additional information to the decoder but whose outputs we do not use (\autoref{sec:aux_factors}).} which help the model keep track of timing. Our main contributions in this paper are thus the following:
\begin{itemize}
    \item We show that target factors can be adapted to predict durations alongside phoneme sequences to jointly optimize translation quality and speech overlap for automatic dubbing.
    \item We design auxiliary counters that further improve the speech overlap by providing extra information to the model to keep track of timing information.
    \item We evaluate our models
    and show that our approach improves upon previous work which instead proposed a model generating interleaved sequences of phonemes and corresponding durations. \todo{human eval?}
    \item We release our 
    implementation\footnote{\textit{https://github.com/awslabs/sockeye/pull/1082}}
    and scripts\footnote{\textit{https://github.com/amazon-science/iwslt-autodub-task}}
    sufficient for replication,
    to enable future research in this area.
\end{itemize}

\section{Related Work}
Standard automatic dubbing methods \cite{federico-etal-2020-speech} usually follow the pipeline where the machine translated transcript is segmented into phrases and pauses via prosodic alignment \cite{oktem19_interspeech,federico2020prosody,virkar2021prosody,virkar2022prosody}, and the final output is synthesized into speech via TTS. 
Since this pipeline can result in output speech needing to be stretched unnaturally in order to satisfy timing constraints, some prior works have tried to avoid the separate prosodic alignment step through training models to predict pauses within translations \cite{Tam2022}, integrating isochrony constraints in MT decoding \cite{saboo-baumann-2019-integration} or by optimizing prosody jointly with the TTS \cite{hu2021neural}.

As a proxy for isochrony, some prior works have proposed optimizing isometry, i.e. generating translations which match the number of characters in the source text \cite{lakew2022isometry,lakew2021verbosity}, but this has been shown to be weakly correlated to isochrony
\cite{brannon2022dubbing}.

Concurrent work \cite{videodubber} 
predicts word durations along with words and presents a novel loss function for decoding. 
They do not elaborate on how word durations are used to generate speech,
and we 
were unable to compare results due to both their translation and TTS implementations being publicly unavailable. %
Other recent work \cite{chronopoulou_jointly_2023}
has presented a simple sequence-to-sequence approach to generate interleaved sequences of phonemes and corresponding duration. We follow the data/model setup and use their approach as our baseline.

Target factors have been used in statistical MT to explicitly model morphology \cite{bojar2007english}. 
They were adopted in neural machine translation to simultaneously translate lemmas with their corresponding parts of speech
\cite{garcia-martinez-etal-2016-factored}, and have also been shown to be effective to predict case markers~\cite{nadejde-etal-2017-predicting}, subword separators \cite{wilken2019novel}, 
capitalization, or gender information \cite{niu2021faithful} 
decoupled from output tokens. 
In the area of isochronous MT, they have been used to predict pause markers as an alternative to generating an explicit token \cite{Tam2022}.

\begin{table*}[t]
    \centering
    \caption{An example target sequence along with its target factors. The corresponding word sequence is shown in the first row but not used in the actual model. The factors are time-shifted internally to condition factor outputs on the main output, which is not shown here. NULL is a padding token used to align the tokens correctly after the internal shift, so that the model sees $f_0^{\mathrm{total}}$, $f_0^{\mathrm{pause}}$, and $f_0^{\mathrm{segment}}$ before generating the first phoneme and duration.}
    \setlength\tabcolsep{4.5pt}
    \begin{tabular}{l|c|ccccc|ccc|ccc|c|ccc}
    \toprule
        Target text & & \multicolumn{5}{c|}{don't} & \multicolumn{3}{c|}{you} & \multicolumn{3}{c|}{know} & [pause] & \multicolumn{3}{c}{it} \\
        $f^{\mathrm{main}}$ & NULL & D & OW1 & N & T & \textlangle eow\textrangle & Y & UW1 & \textlangle eow\textrangle & N & OW1 & \textlangle eow\textrangle & [pause] & IH0 & T & \textlangle eow\textrangle \\
        $f^{\mathrm{dur}}$ & NULL & 2 & 5 & 6 & 8 & 0 & 3 & 7 & 0 & 5 & 41 & 0 & 0 & 5 & 7 & 0 \\
        $f^{\mathrm{total}}$ & 89 & 87 & 82 & 76 & 68 & 68 & 65 & 58 & 58 & 53 & 12 & 12 & 12 & 7 & 0 & 0 \\
        $f^{\mathrm{pause}}$ & 1 & 1 & 1 & 1 & 1 & 1 & 1 & 1 & 1 & 1 & 1 & 1 & 0 & 0 & 0 & 0 \\
        $f^{\mathrm{segment}}$ & 77 & 75 & 70 & 64 & 56 & 56 & 53 & 46 & 46 & 41 & 0 & 0 & 12 & 7 & 0 & 0 \\
        \bottomrule
    \end{tabular}
    \label{tab:factor_example}
\end{table*} 

\section{Method} \label{sec:aux_factors}

We propose predicting phoneme durations as target factors \cite{garcia-martinez-etal-2016-factored},
instead of interleaving phonemes and phoneme durations \cite{chronopoulou_jointly_2023}.
Target factors are additional output layers to produce multiple outputs at each decoder step.
There are separate embedding layers for each target factor, and all the factor embeddings are concatenated to the main target embedding and provided as input to the decoder. To condition the factor outputs upon the main output, factors are shifted such that the factors corresponding to output token $y_t$ are generated at step $t+1$.
We use the Sockeye\footnote{\url{https://github.com/awslabs/sockeye}}
target factor implementation.
In contrast to the interleaved baseline \cite{chronopoulou_jointly_2023}, target factors 
allow us to model the phonemes and durations separately while still ensuring that they are conditioned on each other.
It also significantly decreases the sequence length and eliminates the possibility of producing invalid output (e.g. two durations in a row).

In addition to the main output $f^{\mathrm{main}}$ and corresponding durations being generated as a target factor ($f^{\mathrm{dur}}$), we propose additional input embeddings in the decoder to help the model keep track of the isochrony constraints, which we denote auxiliary counters.
The counters are implemented identically to target factors (i.e. each counter has an embedding layer whose embeddings are concatenated to the target embedding\footnote{As an alternative to trained embeddings for each numeric counter value, 
we also tried fixed sinusoidal embeddings, inspired by the positional embeddings used in transformers \cite{vaswani-transformer}.
We found that models with sinusoidal embeddings converged faster but achieve lower translation quality. It is not clear why sinusoidal embedding would lower translation quality, and we hope to better understand (and perhaps improve on) this in future work.}), 
except that the counters are not predicted at inference. Instead, the values of the counters are 
calculated at each time step based on the prior durations predicted by the model and used as input in the next step.
These counters are:
\begin{itemize}
    \item \textbf{Total frames remaining} ($f_t^{\mathrm{total}}$): Keeps track of the total number of frames remaining in the sentence. This is initialized by the total desired duration of the sentence and is decremented by the phoneme duration at each output step.
    \begin{equation}
        f_t^{\mathrm{total}} = f_{t-1}^{\mathrm{total}} - f_t^{\mathrm{dur}}
    \end{equation}
    \item \textbf{Pauses remaining} ($f_t^{\mathrm{pause}}$): Keeps track of the number of pauses remaining in the sentence.
    \begin{equation}
        f_t^{\mathrm{pause}} =
        \begin{cases}
            f_{t-1}^{\mathrm{pause}} - 1, & \text{if}\ f_{t}^{\mathrm{main}}=\mathrm{[pause]} \\
            f_{t-1}^{\mathrm{pause}}, & \text{otherwise}
        \end{cases}
    \end{equation}
    \item \textbf{Segment frames remaining} ($f_t^{\mathrm{segment}}$): Keeps track of the number of frames remaining in a segment, i.e. until a pause is generated, or the sentence ends. This is initialized by the segment durations from the source sentence, and is decremented by the phoneme duration at each step until a \texttt{[pause]} is generated.
    \begin{equation}
        f_t^{\mathrm{segment}} =
        \begin{cases}
            f_{t-1}^{\mathrm{segment}} - f_t^{\mathrm{dur}}, & \text{if}\ f_{t}^{\mathrm{main}}\neq\mathrm{[pause]} \\
            \text{next segment duration}, & \text{otherwise}
        \end{cases}
    \end{equation}
\end{itemize}

All of these auxiliary counters are calculated from the phonemes and durations in pre-processing for training, and calculated at each time step at inference time. While the model can generate predictions for counters as target factors, we only use the counters to help the model keep track of its state and discard their outputs.\footnote{Additionally, our best models are trained without any gradient coming from the auxiliary counter predictions, effectively removing the part of the network predicting auxiliary counter outputs.}
An example of a target sequence along with its target factor and counters is shown in \autoref{tab:factor_example}.

The implemented behavior of target factored models in Sockeye at inference time is to predict target factors and then feed those predictions back into the model at the next inference step.
For counters (where we are trying to help the model keep track of timing), we found it critical to correctly calculate counter values according to the equations in \autoref{sec:aux_factors} before feeding them back to the decoder at the next time step.
Compared to the default Sockeye behavior for target factors, 
this improved speech overlap significantly 
(from 0.9181 to 0.9972)
without affecting translation quality.

We show in future sections that our method is able to satisfy the duration constraints almost perfectly while maintaining reasonable translation accuracy. 
However, in practice we do not want to achieve perfect speech overlap
because it can result in poor translations or speech that is shortened/lengthened to the point where it sounds \emph{unnatural}. 
In fact, analysis of human dubbing
\cite{brannon2022dubbing} 
has shown that human dubbers prioritize naturalness and translation quality over speech overlap.\footnote{Median overlap is just 0.731 in a large corpus of human dubs.}
For this reason, following prior work in 
isometric MT \cite{wilken-matusov-2022-appteks}
and 
automatic dubbing \cite{chronopoulou_jointly_2023},
we add gaussian noise to the segment durations in our training data. 
This creates training examples where part or all of the translation ends slightly before or after the counters reach zero, and the model learns to be flexible with the timing information.

\section{Experimental Setup} \label{sec:baseline}

We use the English-German subset of CoVoST-2\footnote{\url{https://github.com/facebookresearch/covost}} 
as our dataset, consisting of English audio clips and transcripts along with German text translations. Each clip roughly corresponds to a sentence.
We run the Montreal Forced Aligner (MFA) \cite{mcauliffe17_interspeech} on the English audio and transcripts to get sequences of phonemes with corresponding durations. This sequence becomes our target and the German transcripts are used as the source. 
We mark silence of more than 0.3 seconds with \texttt{[pause]} tokens in the target phoneme sequence in order to be able to reinsert these periods of silence in final dubs. 
We also mark the end of words with \texttt{<eow>} tags. 
We calculate the duration of each segment (speech without pauses) by adding the phoneme durations between pauses, bin them into 100 bins of approximately equal frequency to avoid sparsity, and add these as tags to the source texts. We apply BPE \cite{sennrich-etal-2016-neural} on the German text with 10k merges.
Our final dataset consists of 289,074 training examples, with 15,499 examples in the validation set and 15,413 in the test set.

For all models, we use a standard transformer-base architecture, augmented with target factors and counters where applicable, trained with a maximum batch size of 32768 tokens for 600 epochs, with a dropout probability of 0.3 and label smoothing 0.1. We save checkpoints every 2000 updates and pick the best checkpoint according to COMET on the validation set.

Our baseline follows the approach described by \cite{chronopoulou_jointly_2023}, which is a simple Transformer sequence-to-sequence model. The input is the subword-level source text, with binned segment durations appended as tags to the end of the sequence and the output sequence is an interleaved sequence of phonemes and corresponding durations, with \texttt{<eow>} tags to mark the end of each word and \texttt{[pause]} tokens to mark the end of a segment. As an example, a source sentence is formatted as \texttt{Das weißt du nich@@ t@@ ? <||> <bin4> <bin1>} with the corresponding target sequence \texttt{D 2 OW1 5 N 6 T 8 <eow> Y 3 UW1 7 <eow> N 5 OW1 41 <eow> [pause] IH0 5 T 7 <eow>}.

Additionally, we train a German$\rightarrow$English machine translation model using the same datasets at the subword level (instead of phoneme outputs), and a model to translate German text to English phoneme sequences without durations. These two models act as baselines to measure how much the translation quality deteriorates for models with duration constraints.

Since our models output sequences of phonemes, we train a Transformer seq2seq model on the same dataset to transform English phoneme sequences into sequences of English words. Translation quality is then evaluated using BLEU\footnote{SacreBLEU: BLEU|\#1|c:lc|e:no|tok:none|s:exp|v:2.3.1} \cite{papineni-etal-2002-bleu,post-2018-call}, Prism \cite{thompson-post-2020-automatic, thompson-post-2020-paraphrase}, and COMET\footnote{wmt20-comet-da} \cite{rei-etal-2020-comet}. We find the metrics to be highly correlated in our results, and thus report only BLEU scores.

To quantify speech overlap between the reference (ref.) and the hypothesis (hyp.), we use the relative difference of duration  between reference segments and predicted translated segments, averaged over all segments in the dataset:
\begin{equation}
    \mathrm{Speech\ Overlap} = 1 - \frac{|\mathrm{ref.\ duration} - \mathrm{hyp.\ duration}|}{\mathrm{ref.\ duration}}
\end{equation}

As an additional automatic metric, we also count the number of sentences in the validation and test sets where the wrong number of pauses is generated.

\todo[inline]{describe human evaluation setup (pairwise, MOS scores)}

\begin{table}[t]
    \centering
    \setlength\tabcolsep{4pt}
    \caption{Summary of key results for some representative models on the test set. The models with all the target counters use the optimal configuration from \autoref{sec:factor_params}.}
    \begin{tabular}{lcc}
        \toprule
        Model Configuration & BLEU $\uparrow$ & Speech Overlap $\uparrow$ \\
        \midrule
        Text to text (MT) & 38.0 & -- \\
        Text to phonemes & 35.8 & -- \\
        \midrule
        Interleaved, no noise & 32.0 & 0.8702 \\
        Interleaved, noised 0.2 & 35.4 & 0.7105 \\
        \midrule
        Single target factor, no noise & 33.8 & 0.8931 \\
        \hspace{2mm} + all counters, no noise & 34.0 & 0.9887 \\
        \hspace{2mm} + all counters, noised 0.1 & 35.6 & 0.8649 \\
        \bottomrule
    \end{tabular}
    \label{tab:results_summary}
\end{table}

\section{Analysis} \label{sec:factor_params}

We train models to predict phoneme durations as a target factor and use all the auxiliary counters described in \autoref{sec:aux_factors}.

\begin{figure}[ht]
    \centering
    \includegraphics[width=0.48\textwidth]{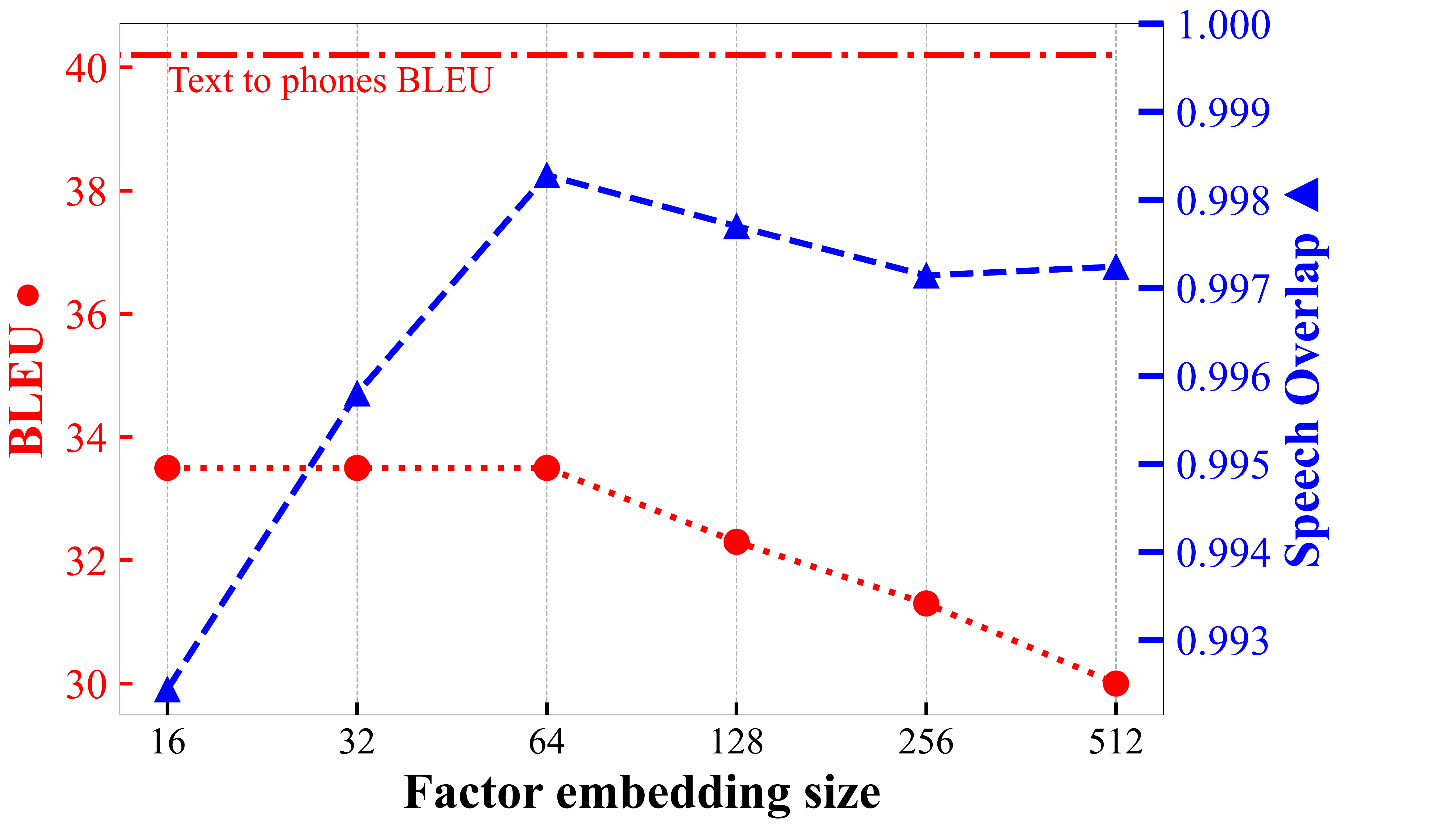}
    \caption{Results on the validation set varying the factor and counter embedding sizes. $f^{\mathrm{pause}}$ embedding size is always half of the other counters. Models trained with equal loss weights on factors and counters, on data with clean segment durations.}
    \label{fig:factor_emb_size}
\end{figure}

\textbf{Embedding Size:} Since the factored architecture adds a large number of parameters in the form of embedding matrices and output layers to the model, we want to optimize the factor embedding size so that it is large enough to adequately represent all the possible factor/counter values while not being too large to train in our limited data scenario. We sweep through a range of embedding sizes (\autoref{fig:factor_emb_size}) and find that 64 dimensions is an optimal size. We set the embedding size for the $f^{\mathrm{pause}}$ counter to half of the other counter embeddings since it has much fewer possible values than the other counters.

\textbf{Counter Loss Weights:} At training time, counters are predicted at each step just like target factors, and all the factors/counters are assigned an equal weight for loss computation by default, i.e. the cross-entropy losses for the output and all target factors are simply summed. However, it is possible to generalize the loss by assigning different weights to the outputs. Since we do not use the outputs for the counters, we can set the weights of the counters to 0, thus letting the model focus on the phoneme and duration outputs that we actually need. We find that zeroing the loss weights of the counters helps improve translation quality by 4.2 BLEU at the cost of a very small drop (0.003) of speech overlap.

\begin{figure}[ht]
    \centering
    \includegraphics[width=0.56\textwidth]{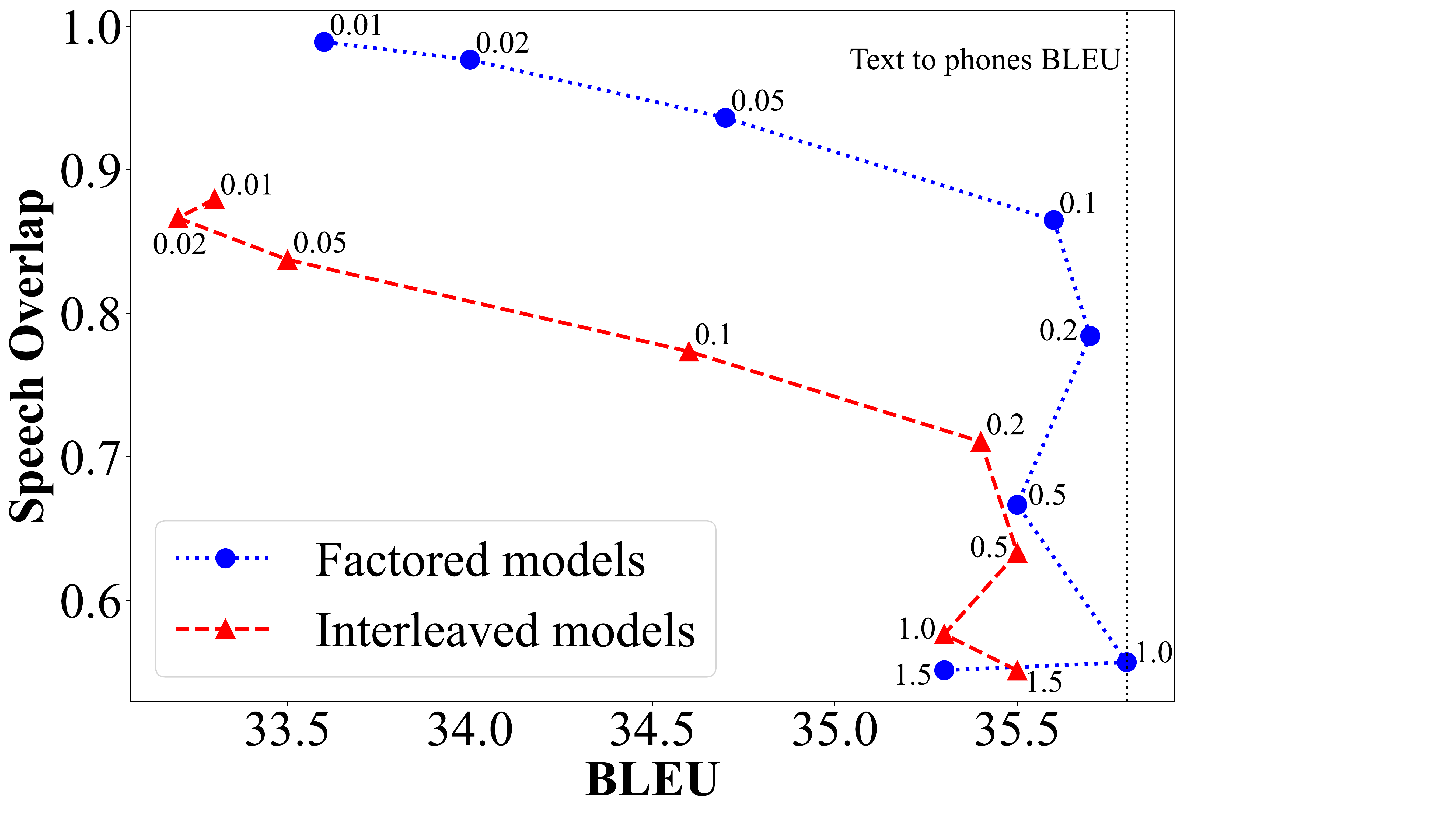}
    \caption{Variation of translation quality (BLEU) and speech overlap with different amounts of noise added to the segment durations. Results shown on the test set. Each point annotation indicates the standard deviation of the added noise.}
    \label{fig:noise_plot}
\end{figure}

\textbf{Adding Noise:} We find that as we add more noise, for both the interleaved as well as factored models, the translation quality increases at the cost of speech overlap, ultimately matching the text-to-phonemes baseline. %
(\autoref{fig:noise_plot}). The amount of noise allows us to control the trade-off between translation quality and speech overlap.

\todo[inline]{However, the primary motivation of adding noise was to avoid unnatural speech, which is very detrimental to user experience. To test this hypothesis, we \dots present human results}

\textbf{Counter Ablations:} To evaluate the effectiveness of the counters and source duration tags, we start with the highest-quality factored model -- embedding sizes 64, 64, 64, 32 with zeroed counter loss weights -- and measure the change in translation accuracy and speech overlap for models with one or more of the counters removed.
From \autoref{tab:ablation}, we can see that removing either the source segment tags or $f^{\mathrm{total}}$ has very little impact on the speech overlap, since the model is able to track the timing information from $f^{\mathrm{segment}}$. 
Removing both the source segment tags and $f^{\mathrm{segment}}$ causes a large drop in speech overlap since the model has no information about segment durations. 
We also see that removing $f^{\mathrm{segment}}$ and $f^{\mathrm{pause}}$ causes a large number of outputs to have the wrong number of pauses.\footnote{We cannot remove only $f^{\mathrm{pause}}$ since $f^{\mathrm{segment}}$ uses $f^{\mathrm{pause}}$ to fetch the correct segment durations.}
These results are consistent with our intuition about the purpose of each of these counters.

\begin{table}[ht]
    \small
    \centering
    \caption{Counter ablation results on the test set. We remove the auxiliary counters and/or source segment durations and measure the effect. W.P. represents the number of sentences in the test set for which the wrong number of pauses is generated.}
    \setlength\tabcolsep{3pt}
    \begin{tabular}{lccc}
        \toprule
        Model configuration & BLEU $\uparrow$ & Overlap $\uparrow$ & W.P. $\downarrow$ \\
        \midrule
        All counters + source durations & 34.0 & 0.9887 & 29 \\
        \midrule
        Without: & & \\
        \hspace{4pt} Source durations & 33.4 & 0.9900 & 25 \\
        \hspace{4pt} $f^{\mathrm{total}}$ & 34.0 & 0.9914 & 8 \\
        \hspace{4pt} $f^{\mathrm{segment}}$ & 34.0 & 0.9258 & 109 \\
        \hspace{4pt} $f^{\mathrm{segment}}$ + $f^{\mathrm{pause}}$ & 33.9 & 0.9294 & 176 \\
        \hspace{4pt} Source durations + $f^{\mathrm{segment}}$ & 34.1 & 0.6214 & 103 \\
        \hspace{4pt} $f^{\mathrm{total}}$ + $f^{\mathrm{segment}}$ & 33.9 & 0.9191 & 20 \\
        \bottomrule
    \end{tabular}
    \label{tab:ablation}
\end{table}

\section{Results}

The translation quality
of the text-to-phones baseline is 2.2 BLEU lower than the text-to-text (i.e.\ standard MT) model (see \autoref{tab:results_summary}).
This is likely due to:
(1) To evaluate the text-to-phoneme model, we are mapping phonemes to words using a seq2seq model and then scoring with word-level metrics. The seq2seq model is not perfect and is likely introducing some errors, making the text-to-phoneme model appear to be worse than it actually is, and 
(2) We used the same parameters for the phoneme model as the text model. 
We did not attempt to optimize the transformer parameters for phonemes, but plan to do so in future work. 

Modeling phoneme durations using target factors improves both translation quality (+1.8 BLEU) and speech overlap (+0.023) relative to the interleaved baseline (see \autoref{tab:results_summary}, no noise settings).

Adding auxiliary counters provides nearly perfect speech overlap (0.9887, perfect score is 1.0) in the no noise setting. It provides substantial improvement in speech overlap (+0.0956) compared to the target factor model without auxiliary counters, while marginally improving translation quality (+0.2 BLEU) (see \autoref{tab:results_summary}, no noise settings).

By adding noise to the speech segment durations, we are able to obtain nearly the same translation quality as the text-to-phoneme  model (35.6 vs 35.8) while still achieving very high speech overlap (0.8649, higher than observed in human dubs).
\todo[inline]{describe impact of noise on speech naturalness}

\section{Qualitative Perception Results}

We intended to perform human evaluation of dubbed videos using crowd source workers but a pilot showed very noisy results, with annotators often appearing to ignore annotations guidelines. We believe this is due at least in part to the large number of factors that affect perception of a dubbed video, including (but not limited to) translation quality, speech quality / naturalness, isochrony, and lip sync. 

We present instead some qualitative conclusions drawn by the authors after watching/listening to many samples. The baseline tends to be the most natural sounding, but the lack of isochrony is disconcerting.\footnote{We believe the lack of isochrony would be even more jarring when viewing dubbed content with multiple speakers.} The proposed models with little or no noise added have much better isochrony, as expected, but often sound a little more robotic than the baseline, and it is not unusual to have a word at the end of a segment repeated (presumably this happens when the translation model finishes a translation but the counters tell the model it must keep producing output). The proposed models with large amounts of noise also sounded a bit unnatural, but for a very different reason. The speech in the test set appears to be fairly slow compared to the training data, while the model produces speech with speaking rates similar to the training data, resulting in speech segments which are often short, resulting in long, often unnatural pauses between speech segments. The proposed models with with noise of around 0.1 seem to be the best compromise between isochrony and naturalness/translation quality, consistent with the automatic evaluation (see \autoref{fig:noise_plot}).

\section{Conclusions}
In this paper, we have shown that target factors can be used to predict phoneme durations alongside translated phoneme sequences to jointly optimize translation and timing for automatic dubbing. We train models with target factors for duration prediction as well as other auxiliary counters to further guide the model. Automatic evaluation
show that our models out-perform a baseline of training a model to generate interleaved phoneme and duration sequences. \todo{and human?}

\todo[inline]{summarize human results}

\bibliographystyle{IEEEtran}
\bibliography{anthology,custom}

\end{document}